\documentclass[letterpaper]{article} 
\usepackage{aaai25}  
\usepackage{times}  
\usepackage{helvet}  
\usepackage{courier}  
\usepackage[hyphens]{url}  
\usepackage{graphicx} 
\usepackage{natbib}  
\usepackage{caption} 
\usepackage{algorithm}
\usepackage{algorithmic}
\usepackage{tabularx}

\usepackage{newfloat}
\usepackage{listings}
\DeclareCaptionStyle{ruled}{labelfont=normalfont,labelsep=colon,strut=off} 
\floatstyle{ruled}
\newfloat{listing}{tb}{lst}{}
\floatname{listing}{Listing}

\usepackage{amssymb}
\usepackage{amsthm}
\usepackage{lineno}
\usepackage{graphicx}
\usepackage[caption=false,font=footnotesize]{subfig}
\usepackage{amsmath}
\usepackage{pifont}
\usepackage{makecell}
\usepackage{array}
\usepackage{booktabs}
\usepackage[figuresright]{rotating}
\usepackage{multirow}
\usepackage{makecell}
\usepackage{url}
\usepackage{bbding}
\usepackage{wasysym}
\usepackage{makecell}
\usepackage{soul}
\usepackage{color, xcolor}
\soulregister\cite7
\soulregister\eqref7
\def\etal{\emph{et al. }}
\definecolor{tab-best}{RGB}{163, 212, 184}
\definecolor{tab-second}{RGB}{156, 195, 229}

\title{GRPose: Learning Graph Relations for Human Image Generation\\ with Pose Priors}

\author {
    Xiangchen Yin\textsuperscript{\rm 1,5},
    Donglin Di\textsuperscript{\rm 2},
    Lei Fan\textsuperscript{\rm 3},
    Hao Li\textsuperscript{\rm 2},
    Wei Chen\textsuperscript{\rm 2*},\\
    Xiaofei Gou\textsuperscript{\rm 2},
    Yang Song\textsuperscript{\rm 3},
    Xiao Sun\textsuperscript{\rm 4},
    Xun Yang\textsuperscript{\rm 1*}
}
\affiliations {
    \textsuperscript{\rm 1}University of Science and Technology of China \ 
    \textsuperscript{\rm 2}Space AI, Li Auto \\
    \textsuperscript{\rm 3}University of New South Wales \
    \textsuperscript{\rm 4} Hefei University of Technology \\
    \textsuperscript{\rm 5} Institute of Artificial Intelligence, Hefei Comprehensive National Science Center \\
    yinxiangchen@mail.ustc.edu.cn \ \{didonglin,lihao43,chenwei10,gouxiaofei\}@lixiang.com\\
    \{lei.fan1,yang.song1\}@unsw.edu.au \
    sunx@hfut.edu.cn \ xyang21@ustc.edu.cn \\
    
}

\usepackage{bibentry}

\begin{document}

\maketitle

\def\thefootnote{*}\footnotetext{\noindent Wei Chen and Xun Yang are co-corresponding authors. The work was done within intern at Li Auto. }\def\thefootnote{\arabic{footnote}}

\begin{abstract}

Recent methods using diffusion models have made significant progress in human image generation with various control signals such as pose priors. However, existing efforts are still struggling to generate high-quality images with consistent pose alignment, resulting in unsatisfactory output. In this paper, we propose a framework that delves into the graph relations of pose priors to provide control information for human image generation. The main idea is to establish a graph topological structure between the pose priors and latent representation of diffusion models to capture the intrinsic associations between different pose parts. A Progressive Graph Integrator (PGI) is designed to learn the spatial relationships of the pose priors with the graph structure, adopting a hierarchical strategy within an Adapter to gradually propagate information across different pose parts. Besides, a pose perception loss is introduced based on a pretrained pose estimation network to minimize the pose differences. Extensive qualitative and quantitative experiments conducted on the Human-Art and LAION-Human datasets clearly demonstrate that our model can achieve significant performance improvement over the latest benchmark models. The code is available at \url{https://xiangchenyin.github.io/GRPose/}.

\end{abstract}

%

\section{Introduction}

\begin{figure*}[t]
\centering
    \includegraphics[width=0.95\textwidth]{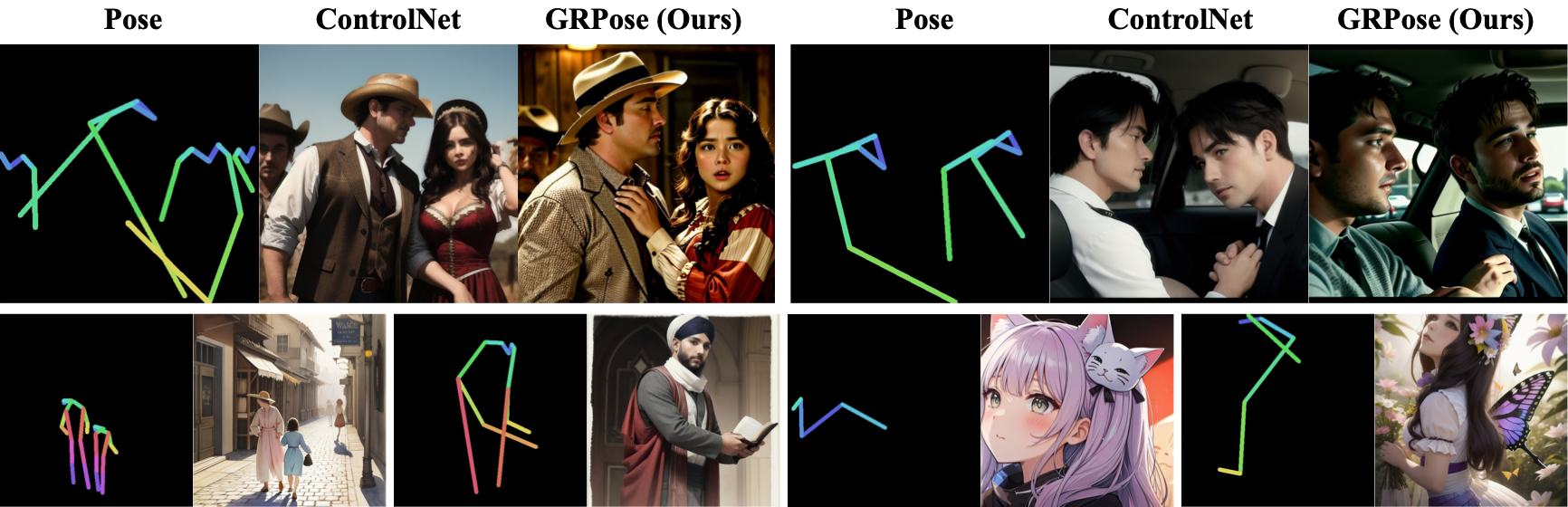}
    \captionof{figure}{Examples of the pose-guided human image generation task.
    The first row illustrates the generated results of the ControlNet and our GRPose methods while the second row visualizes the pose alignment across different base models. GRPose generates better results by well aligning with pose prior and scaling outputs to $512 \times 512$ pixels. } \label{Fig1}
\end{figure*}

Human image generation aims to synthesize high-quality images under specific conditions based on a series of prompts, such as canny edge, pose and depth~\cite{yang2023diffusion,wang2024eulermormer}. Its diverse applications range from animation~\cite{corona2024vlogger} and game production~\cite{pan2024synthesizing} to other fields. Early methods~\cite{men2020controllable, ma2017pose} primarily adopted variational autoencoders (VAEs)~\cite{vae} or Generative Adversarial Networks (GANs)~\cite{goodfellow2020generative}, leveraging a source image to synthesize target images with specific human attributes. Although these methods achieve control through the reference appearance, the synthesis process is unstable and the training heavily depends on the distribution of the source images. Recently, Stable Diffusion (SD)~\cite{ldm} and its variants~\cite{sdxl} have been developed to address these limitations, in which high-fidelity human images are synthesized with the help of a prompt. Considering the data availability and computation costs, controllable diffusion models~\cite{controlnet, t2iadapter} further introduce a learnable control branch into the frozen SD model, enabling spatial control of the generative results based on the  provided conditions, \textit{e.g.,} depth maps and segmentation masks.  

In recent years, ControlNet (Zhang et al. \citeyear{controlnet}) has emerged as a new benchmark by fine-tuning the frozen SD model with trainable copy parameters, enabling spatial control through conditional inputs. HumanSD~\cite{humansd} adopts a heat-guided loss to achieve pose control. While these methods synthesize images with satisfactory semantics and style, they are still struggling to produce high-quality output based on the pose priors, often resulting in unrealistic body alignment (as shown in Figure~\ref{Fig1}). A key challenge in existing approaches is the poor alignment with the given pose priors. We observed that these methods incorporate human pose information into frozen generative models based on Euclidean space, which inadequately models the nonlinear and higher-order relationships between different parts of pose priors, particularly in terms of joint connections and overall coordination.


In this study, we propose to capture the topological relationships among different pose parts by leveraging a graph structure to extract intrinsic associations. We propose a framework named Graph Relation Pose (GRPose) to guide the process of stable diffusion. We follow the ControlNet framework that utilizes both a text prompt and a pose prior to generate human images, and primarily fine-tune the Adapter which is a set of trainable copy parameters. We aim to establish a graph topological structure on pose priors and latent representations, adopting Graph Convolutional Networks (GCNs) (Kipf et al. \citeyear{kipf2016semi}; Han et al. \citeyear{han2022vision}) to effectively capture the higher-order relationships between different pose parts. We introduce a graph construction mechanism called the Progressive Graph Integrator (PGI). PGI treats each spatial part of pose priors and latent representations as graph nodes, mapping the spatial relationships of the pose into a graph structure. This mechanism distills pose information into each layer within the Adapter through a hierarchical structure, allowing gradual refinement of the pose while preserving the authenticity of the appearance. Additionally, we propose a novel pose perception loss, which adopts a pretrained pose estimation network to minimize the pose differences between outputs and original images, further encouraging alignment of the synthesized output with pose priors. 
Our contributions are summarized as follows:
\begin{itemize}
\item We propose a novel framework, termed GRPose, that exploits a graph structure to effectively capture the topological and spatial information of pose priors for human image generation, offering a new perspective in the field.
\item We design a Progressive Graph Integrator (PGI) to explicitly capture the intrinsic associations between different pose parts, distilling pose information into each layer.
\item We introduce a novel pose perception loss that utilizes a 
pose estimation network to minimize pose differences.
\item We conduct extensive 
experiments on the Human-Art~\cite{ju2023human} and LAION-Human datasets~\cite{ju2023humansd}. Our GRPose achieves superior performance compared with advanced methods across multiple  metrics, particularly in terms of pose guidance alignment.

\end{itemize}

\begin{figure*}[t]
\centerline{\includegraphics[width=1.0\linewidth]{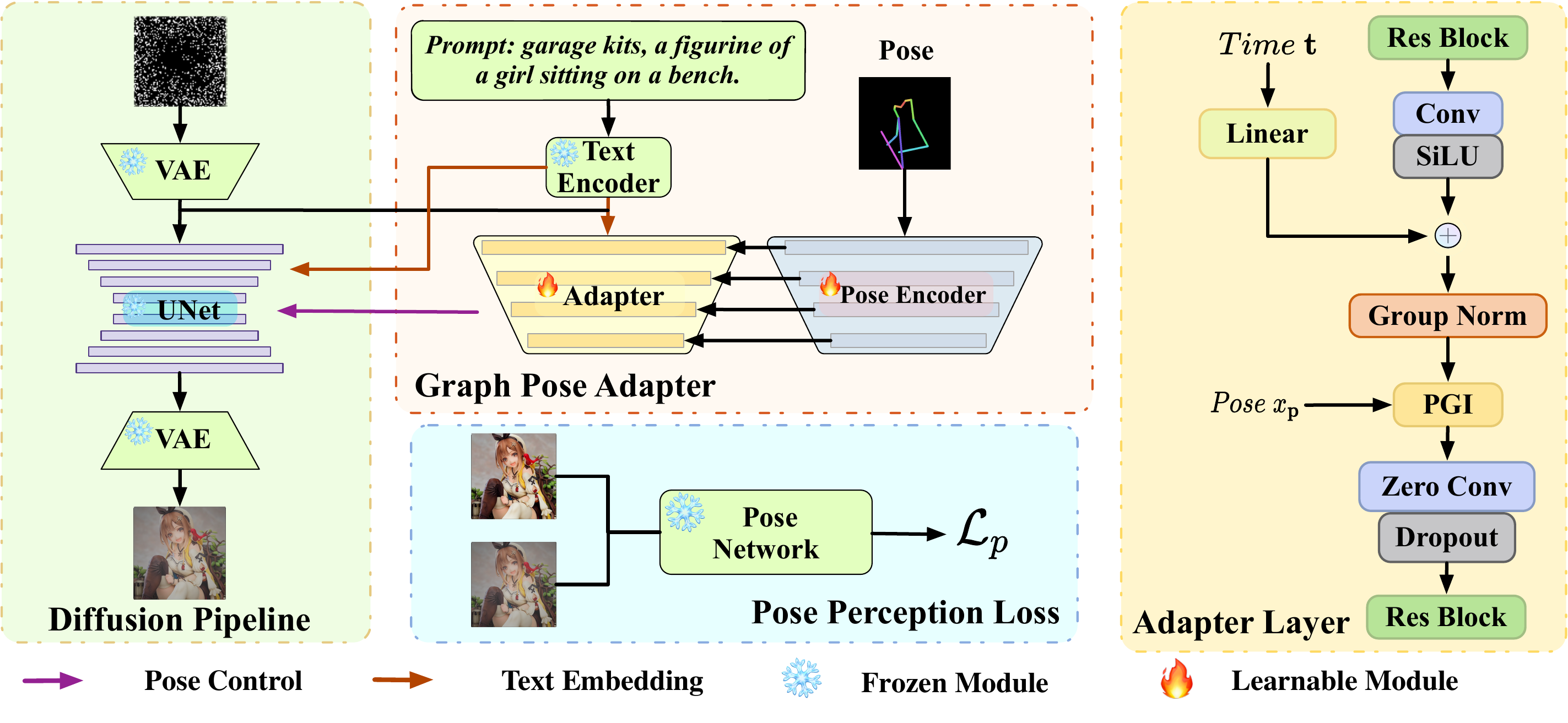}}
\caption{Overview of Graph Relation Pose (GRPose). The Pose Encoder is adopted to capture multi-level scales of pose priors within a hierarchical structure, where a Progressive Graph Integrator is incorporated to capture graph relationships between different pose parts. The Pose Perception Loss adopts a pre-trained pose estimation network to regularize the pose alignment.} \label{Overview}
\end{figure*}

\section{Related Work}

\subsection{Pose-Guided Human Image Generation}

Traditional pose-guided human generation methods take a specific source image and pose condition as input to generate images that retain the appearance of the source image while presenting a specified pose. Recently, deep learning techniques have achieved significant progress in various domains \cite{fan2022fast, liang2024tocoad}. Some studies~\cite{lv2021learning, ma2021must, yang2021towards} are based on GANs or VAEs to convert the task as a conditional generation. Bhunia \etal~\cite{bhunia2023person} leveraged a Person Image Diffusion Model (PIDM) to synthesize images by learning a noise distribution and proposed a cross-attention-based Texture Diffusion Module (TDB) to align the relationship between appearance and pose information. Zhang \etal~\cite{zhang2022exploring} proposed a dual-task pose transformer and introduced an auxiliary task, connecting different branches to obtain correlation between features by building attitude conversion modules. Shen \etal~\cite{shen2023advancing} introduced a Progressive Conditional Diffusion Model (PCDM), which incrementally bridges the gap between character images under target poses and source poses through a three-stage process. HumanSD~\cite{humansd} finetuned the diffusion model with a pose-guided heatmap loss and created a new LAION-Human dataset. Stable-Pose~\cite{wang2024stable} introduced a coarse-to-fine attention masking strategy into ViT to obtain more accurate pose guidance. Despite the significant progress of existing methods, challenges remain in handling complex poses.

\subsection{Controllable T2I Diffusion Model}

Large Diffusion models~\cite{ldm, ramesh2022hierarchical} generate high-quality images under a set of prompts. Ho \etal~\cite{ho2022classifier} proposed classifier-free guidance that combines conditional prediction with unconditional prediction. ControlNet~\cite{controlnet} introduces a trainable control branch by incorporating a weight copy of the diffusion model, allowing spatial control of generative images using depth maps, segmentation masks and more. T2I-Adapter~\cite{t2iadapter} trains a lightweight adapter to guide the frozen SD model. UniControlNet~\cite{unicontrolnet} utilizes two additional control modules to achieve control over multiple conditions. Although these methods demonstrate strong control capabilities, they struggle with handling complex pose relationships.

\section{Methodology}

\subsection{Overview of Our GRPose}

Our aim is to generate high-quality human images conditioned on pose priors. Our proposed GRPose consists of three main components: Diffusion Pipeline, Graph Pose Adapter and Pose Perception Loss, as shown in Figure~\ref{Overview}. Given a pose condition $c_p \in \mathbb{R}^{H\times W \times C}$ and a text prompt as inputs, the CLIP text encoder~\cite{clip} converts the text into its embedding $c_t$. In the Pose-Guided Diffusion Pipeline, we use Stable Diffusion~\cite{ldm} as the base diffusion model, which includes a VAE for image encoding and decoding, and a U-Net for noise estimation. Specifically, an image is fed into the encoder of the VAE to obtain the latent representation $z_0$. The optimization objective is to maximize the logarithmic likelihood of the data distribution, as defined:
\begin{equation}
\mathcal{L}_d=E_{\mathbf{z}_{t}, t, \epsilon_{t} \sim \mathcal{N}(\mathbf{0}, \mathbf{I})}\left[\left\|\epsilon_{t}-\epsilon_{\theta}\left(\mathbf{z}_{t}, \mathbf{c}_{t}, \mathbf{c}_{p}, t\right)\right\|^{2}\right],
\end{equation} 
where $z_t$ denotes the latent representation at time step $t$, $\mathcal{N}$ denotes the standard normal distribution, $\epsilon_{\theta}$ denotes the noise prediction network and $\epsilon_{t}$ denotes real noise.

Within the entire framework, the Graph Pose Adapter is a trainable component that encodes the pose condition $c_p$ into a graph structure and integrates it into the Adapter through a hierarchical structure. At the beginning of each encoder layer in the Adapter, the encoded pose and the current latent representation are fed into the Progressive Graph Integrator (PGI) to capture the topological relationships between different pose parts through graph learning. This process fine-tunes the Adapter to transfer the control signals into the SD model, producing the synthesized image $\hat{x}$. Additionally, to further encourage the alignment of synthesized output with pose priors, the pose perception loss is formulated using a pretrained pose estimation network to quantify the pose differences between the outputs and original images.

\subsection{Progressive Graph Integrator}

\begin{figure}[t]
\centerline{\includegraphics[width=1.0\linewidth]{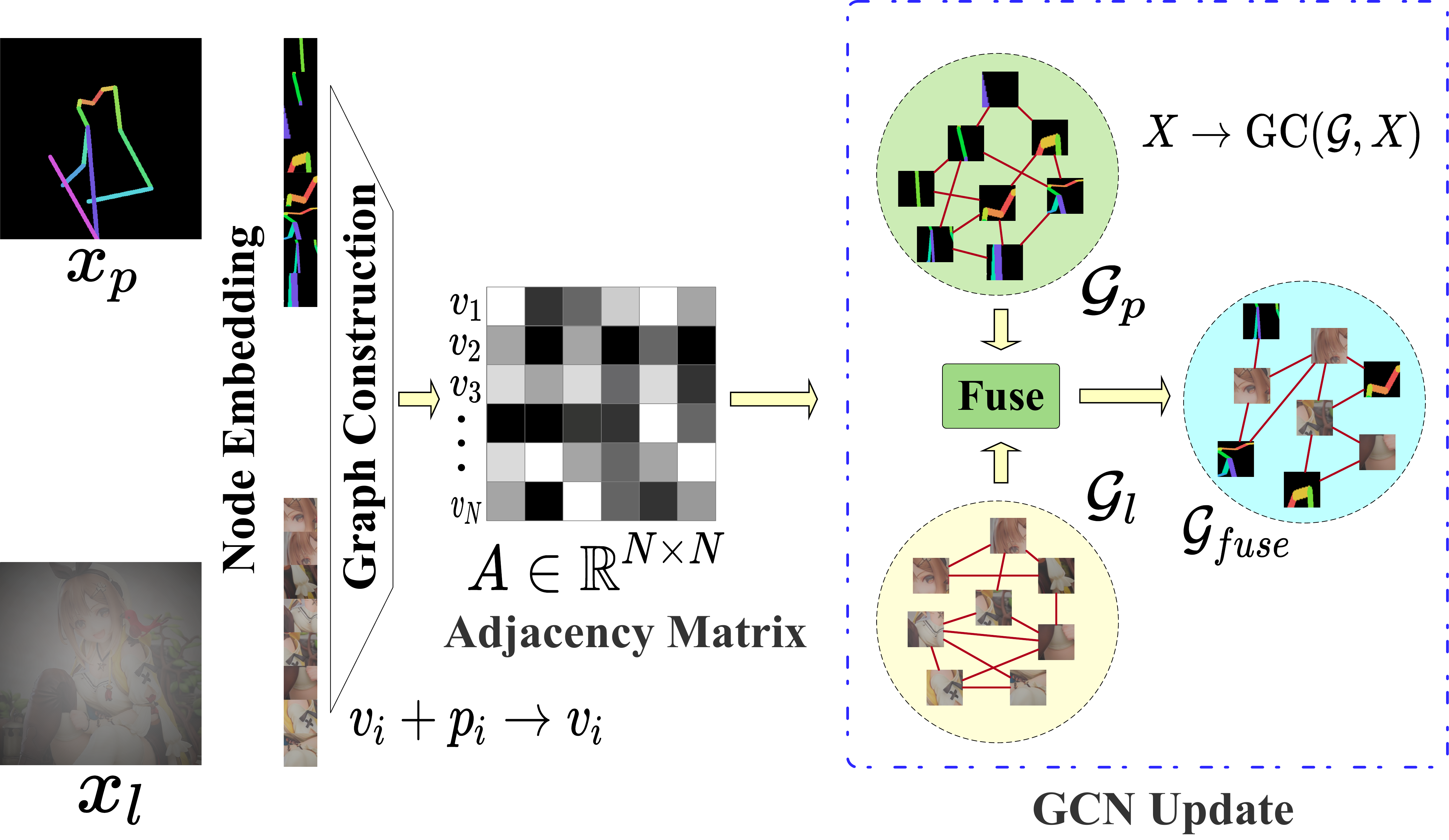}}
\caption{Details of Progressive Graph Integrator (PGI). The pose prior $x_p$ and latent representation $x_l$ are gridded to construct graphs $\mathcal{G}_p$ and $\mathcal{G}_l$ respectively, where GCNs are employed to fuse and update the information.} \label{PGI}
\end{figure}

We adopt a pose encoder with $L$ layers ($L$=4, similar to the Adpater) to capture multi-level scales of pose priors within a hierarchical structure, where a PGI is incorporated into each encoding layer and its corresponding layer in the Adapter. In the PGI, each spatial point in the encoded pose prior $x_p$ and the latent representation $x_l \in \mathbb{R}^{H\times W \times C}$ is treated as a node to construct the graph structures, as shown in Figure~\ref{PGI}. By doing this, both local features of spatial relationships in the pose prior and latent representation are captured. Then, we proceed to construct the feature graph $\mathcal{G} = (\mathcal{V}, \mathcal{E})$, where the node set is defined as $\mathcal{V} = \{v_1, v_2, \dots, v_N\} $ where $ N = H \times W $ denotes the number of nodes. The graph edges between nodes $v_i $ and $ v_j $ are established by applying a \textit{K}-nearest neighbors ($K$NN) search algorithm:
\begin{equation}
N_K(v_i) = \{v_j \in V \mid d(v_i, v_j) \leq K\},
\end{equation}
where $d$ denotes the $L_2$ distance between two node vectors. Notably, we incorporate positional encoding $p_i$ into each node ( $v_i \leftarrow v_i + p_i $), where $p_i $ denotes the spatial positions of the corresponding patch. A set of edge $\mathcal{E}$ can be obtained as:
\begin{equation}
\mathcal{E}=\{ (v_i, v_j) | v_j \in N_K(v_i), i \neq j \},
\end{equation}
Then, two separate graph structures, $\mathcal{G}_p$ and $\mathcal{G}_l$, are formed to represent the pose prior $x_p$ and latent representastion $x_l$ respectively.  
With these two graphs, we employ graph convolution (GC) layers to facilitate the association of features between nodes within each graph. The GC layer is divided into two steps: aggregation and update. In the aggregation step, features of neighboring nodes are gathered as follows:
\begin{equation}
F_{agg} = \sum_{j \in \mathcal{K}(i)} \left( \mathcal{A}_{ij} \cdot X_j \right),
\end{equation}
where $\mathcal{K}(i)$ denotes the set of the neighboring nodes of the node $i$ determined by the $K$NN algorithm, $\mathcal{A} \in \mathbb{R}^{N\times N}$ is an adjacency matrix, and $X$ denotes the features of nodes. In the update step, the current node representation is updated based on the aggregated features, as follows:
\begin{equation}
X'_i = \boldsymbol{\phi}\left( \boldsymbol{\Theta} \cdot X_i \oplus (1 - \boldsymbol{\Theta}) \cdot F_{agg} \right),
\end{equation}
where $\boldsymbol{\phi}$ denotes the activation function, $\boldsymbol{\Theta}$ is a learnable parameter. Then, the adjacency matrix $\mathcal{A}$ is normalized as:
\begin{equation}
\hat{\mathcal{A}} = D^{-\frac{1}{2}} (\mathcal{A}+I) D^{-\frac{1}{2}},
\end{equation}
where $D$ denotes the diagonal degree of $\mathcal{A}$, and $I$ denotes the identity matrix. The GC process can be described as:
\begin{equation}
\hat{X} = \boldsymbol{\phi} \left(\hat{\mathcal{A}} X W \right) + X,
\end{equation}
where $W$ denotes the weight matrix. The GC layer distills information across different parts for both pose prior and latent representation. Take the pose prior as an example, this process is summarized as:
\begin{equation}
\hat{X_p} = \mathrm{GC}(\mathcal{G}_p, X_p),
\end{equation}
We further fuse the graph features of pose and latent through a fusion layer, which captures the cross-modal interactions. The fusion layer consists of three convolution blocks. The fused features are then used to construct a graph $\mathcal{G}_{fuse}$, which is fed into an additional GC layer to capture the complex associations to refine the feature representation.

\subsection{Pose Perception Loss}

We introduce a novel Pose Perception Loss to enhance the pose alignment during the image synthesis process under the guidance of a pose perception network. Our approach employs a pre-trained pose estimation network, specifically designed to accurately identify and comprehend the poses of individuals within images. By encoding both the generated image $\hat{x}$ and the original image $x$, the discrepancies between them can be captured effectively, defined as follows:
\begin{equation}
\mathcal{L}_{p}= \frac{1}{hw} \sum_{i=0}^h \sum_{j=0}^w \left\|\varphi_p\left(\hat{x}_{ij} \right)-\varphi_p\left(x_{ij} \right)\right\|_{2}^{2},
\end{equation}
where $\varphi_p$ denotes the encoder of the pose estimation network, which extract the features from the pose backbone, $h$ and $w$ denotes the width and height of both $x$ and $\hat{x}$ respectively and $\left\| \cdot \right\|_2$ denotes the Euclidean distance. By directly minimizing the pose differences in the feature space, our method is capable of generating images that are not only visually more realistic but also consistent with pose priors. This loss function can be integrated as a plug-and-play component into existing image generation frameworks, providing additional pose guidance to these models.  The optimization objective of the training process is formulated as:
\begin{equation}
\mathcal{L} = \mathcal{L}_d + \alpha * \mathcal{L}_{p}.
\end{equation}
where $\mathcal{L}_d$ denotes the diffusion loss, and $\alpha$ denotes the weight of the pose perception loss $\mathcal{L}_{p}$. The parameters of the Adapter are updated through this loss function.

\tabcolsep=0.18cm
\begin{table*}[h]
\small
    \caption{Results on the Human-Art and LAION-Human datasets. The \colorbox{tab-best!50}{best results} and the \colorbox{tab-second!50}{second best results} are marked in green and blue, respectively. Results marked with asterisk (*) are reimplemented based on the released models. }
    \centering
    \begin{tabular}{p{0.1\linewidth}ccccccccc}
    \toprule
    
    Dataset & Methods & AP (\%) $\uparrow$ & SAP (\%) $\uparrow$ & PCE $\downarrow$ & FID $\downarrow$ & KID $\downarrow$ & CLIP-Score $\uparrow$ \\
    \midrule
    \multirow{7}{*}{Human-Art} & SD*~\cite{ldm}  & 0.24 & 55.71 & 2.30 & 11.53 & 3.36 & \colorbox{tab-best!50}{33.33} \\
    & T2I-Adapter~\cite{t2iadapter}  & 27.22 & 65.65 & 1.75 & 11.92 & 2.73 & \colorbox{tab-second!50}{33.27} \\
    & ControlNet~(Zhang et al. \citeyear{controlnet}) & 39.52 & 69.19 & 1.54 & \colorbox{tab-second!50}{11.01} &\colorbox{tab-best!50}{2.23} & 32.65 \\
    & Uni-ControlNet~\cite{unicontrolnet}  & 41.94 & 69.32 & 1.48 & 14.63 & \colorbox{tab-second!50}{2.30} & 32.51 \\
    & HumanSD~\cite{humansd}  & \colorbox{tab-second!50}{44.57} & \colorbox{tab-second!50}{69.68} & \colorbox{tab-best!50}{1.37} & \colorbox{tab-best!50}{10.03} & 2.70 & 32.24 

    \\\cmidrule{2-8}
    & GRPose (Ours) &\colorbox{tab-best!50}{49.50} & \colorbox{tab-best!50}{70.84} & \colorbox{tab-second!50}{1.43} & 13.76 & 2.53 & 32.31 \\

    \midrule
    \multirow{7}{*}{LAION-Human} & SD*~\cite{ldm}  & 0.73 & 44.47 & 2.45 & \colorbox{tab-best!50}{4.53} & 4.80 & \colorbox{tab-second!50}{32.32} \\
    & T2I-Adapter*~\cite{t2iadapter}  & 36.65 & 63.64 & 1.62 & 6.77 & 5.44 & 32.30 \\
    & ControlNet*~(Zhang et al. \citeyear{controlnet})  & 44.90 & \colorbox{tab-second!50}{66.74} & 1.55 & 7.53 & 6.53 & 32.31 \\
    & Uni-ControlNet~\cite{unicontrolnet}  & 50.83 & 66.16 & 1.41 & 6.82 & \colorbox{tab-best!50}{4.52} & \colorbox{tab-best!50}{32.39} \\
    & HumanSD~\cite{humansd}  & \colorbox{tab-second!50}{50.95} & 65.84 & \colorbox{tab-best!50}{1.25} & \colorbox{tab-second!50}{5.62} & 7.48 & 30.85 

    \\\cmidrule{2-8}
    & GRPose (Ours)  & \colorbox{tab-best!50}{57.01} & \colorbox{tab-best!50}{67.20} & \colorbox{tab-second!50}{1.29} & 6.52 & \colorbox{tab-second!50}{4.65} & 32.12 \\
    \bottomrule
    \end{tabular}
    \label{tab:comparison}
\end{table*}

\section{Experiments}

\subsection{Experimental Settings}
\textbf{Datasets}. We evaluated our model on the Human-Art~\cite{ju2023human} and LAION-Human~\cite{ju2023humansd} datasets. The Human-Art dataset comprises 50,000 high-quality images from 5 real-world and 15 virtual scenarios, featuring human bounding boxes, key points and textual descriptions. The LAION-Human dataset consists of approximately 1 million image-caption pairs, filtered by high image quality and human estimation confidence scores, using a diverse range of human activities and more realistic images. In the LAION-Human dataset, we randomly selected 200,000 samples for training and 20,000 samples for testing.

\noindent{\textbf{Implementation Details}}. For a fair comparison, our diffusion pipeline adopts the same Stable Diffusion 1.5\footnote{\url{https://huggingface.co/runwayml/stable-diffusion-v1-5}} as previous methods. We used the Adam optimizer~\cite{Kingma_Ba_2014} with an initial learning rate of $1 \times 10^{-5}$. We trained models on $8 \times$ NVIDIA L40S-48GB, using Pytorch and PytorchLightning\footnote{\url{https://lightning.ai/docs/pytorch/stable/}}. The batch size was set as 6, and the number of epochs was 20. The weight of pose loss $\alpha$ was set to 0.01 and we adopted a gradual constraint strategy, adding pose perception loss in the last 5 epochs. We replaced the text prompts with an empty string with a probability of 0.5. For the PGI, we set the parameter $K$ of the KNN algorithm to 9. During inference, the number of steps $T$ of the DDIM sampler was set to 50. We used the MMpose\footnote{\url{https://github.com/open-mmlab/mmpose}} framework as the pre-trained pose estimation networks. It is worth noting that in the evaluation phase we used a different pre-trained pose estimation network during the inference process to assess the accuracy of the pose in the generated images. This ensures the independence and reliability of the evaluation.

\noindent{\textbf{Metrics}}. We adopted Average Precision (AP), Similarity Average Precision (SAP) and Person Count Error (PCE)~(Cheong et al. \citeyear{cheong2022kpe}). Higher scores in AP and SAP indicate better pose alignment. Furthermore, we used the Fréchet Inception Distance (FID)~\cite{FID} and the Kernel Inception Distance (KID)~\cite{KID} to assess the quality of the generated images and used the CLIP-Score~\cite{clip} to assess the alignment between the image and text.

\begin{figure}[t]
\centerline{\includegraphics[width=1.0
\linewidth]{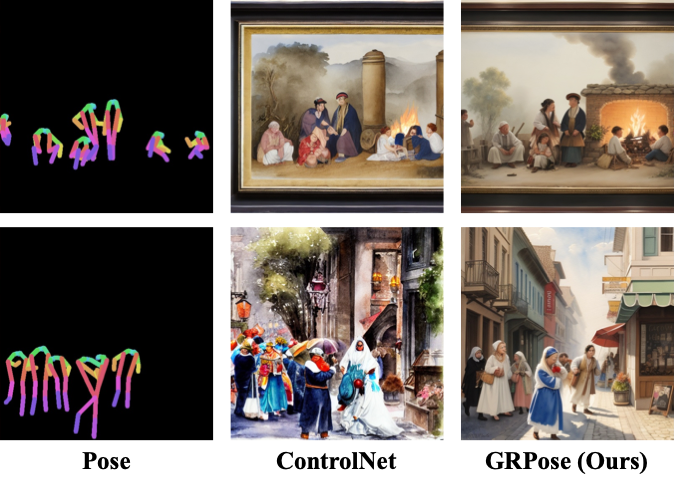}}
\caption{{Two cases of Multi-Pose Generation}. Our model outperforms ControlNet in generating multiple poses. } \label{multi-pose}
\end{figure}

\subsection{Comparison with SOTA Methods}
Quantitative results comparing our method with other state-of-the-art (SOTA) approaches are shown in Table~\ref{tab:comparison}. The experimental results indicate that our GRPose achieved the highest AP and SAP. On the Human-Art dataset, our model attained AP and SAP of 49.50\% and 70.84\% respectively, bringing an improvement of 9.98\% in AP, compared with ControlNet. Similarly, on the LAION-Human dataset, our model showed a consistent improvement of 6.06\% in AP compared to HumanSD~\cite{humansd}. Notably, our method significantly improved pose alignment accuracy, as evidenced by the AP, SAP, and PCE metrics. KID is multiplied by 100 for Human-Art and 1000 for LAION-Human for readability. We observed that SD1.5 using only a text prompt without pose conditions lacked the ability for pose alignment. Although controllable diffusion models such as ControlNet and T2I-Adapter used pose conditions for spatial control, their capabilities of pose alignment were limited due to incorporating it without considering the intrinsic high-order associations between different pose parts. While retaining pose alignment, our model also maintained good performance in image quality and text alignment with minor decline in scores.

\begin{figure}[t]
\centerline{\includegraphics[width=1.0
\linewidth]{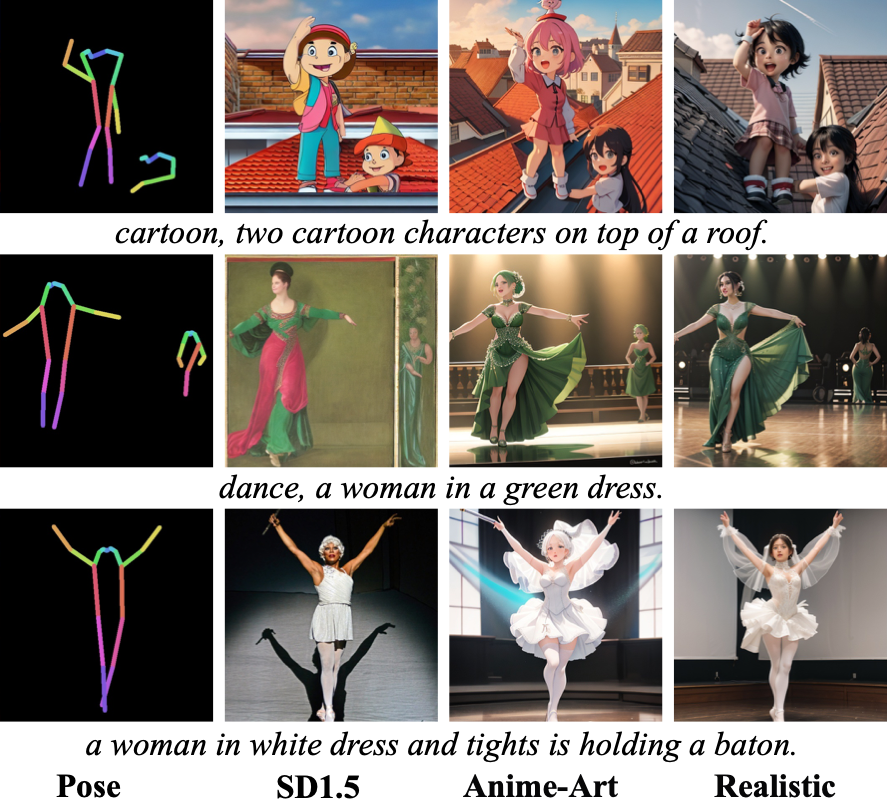}}
\caption{{Qualitative Results of our GRPose with different base diffusion models.} We compared SD1.5, Anime Art and Realistic models of different styles. } \label{Transfer}
\end{figure}

We further compared the visualized qualitative results with other methods~(Zhang et al. \citeyear{controlnet}; Mou et al. \citeyear{t2iadapter}; Ju et al. \citeyear{humansd}; Mou et al. \citeyear{t2iadapter}; Rombach et al. \citeyear{ldm}) 
across several samples, as shown in Figure~\ref{Compa}. We found that SD1.5 lacked effective pose alignment capability since it relies solely on text prompts. Although T2I-Adapter and ControlNet exhibited a good ability to comprehend textual semantics, their performance of pose alignment was poor, and HumanSD also performs poorly in pose alignment. In contrast, our GRPose demonstrated good pose alignment, significantly enhancing the details in human image generation. We also verified the model’s ability in handling multiple poses in a single condition, as shown in Figure~\ref{multi-pose}. Compared with ControlNet, our model can achieve accurate alignment with multiple poses, proving the feasibility of the graph learning approach. Our model addressed issues such as unnatural body positions and misaligned pose, significantly enhancing the accuracy of pose estimation. We perform transfer verification of the base model across different variants of SD-1.5 from the open-source community Civitai, as shown in Figure~\ref{Transfer}. Our GRPose, when integrated with different base models, can effectively improve the performance of pose-guided alignment, demonstrating that GRPose is a plug-and-play and effective solution.

\begin{figure}[t]
\centerline{\includegraphics[width=1.0
\linewidth]{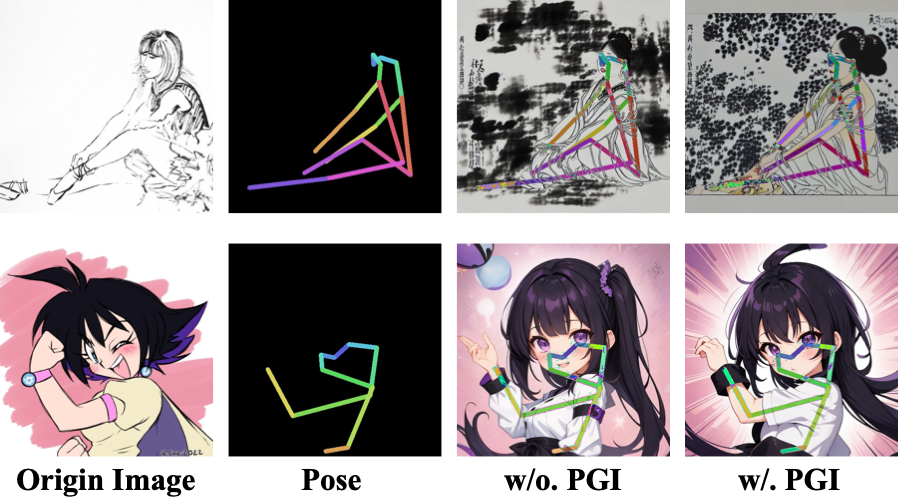}}
\caption{\textbf{Effect of PGI}. Our model with PGI demonstrates better pose guidance due to graph associations.} \label{PGI_Com}
\end{figure}

\begin{table}[t]
    \centering
    \caption{\textbf{Results of each component.} $\mathcal{L}_p$ denotes the Pose Perception Loss. The results verify the effectiveness of each component in our framework. }
    \label{tab:ablation}
    \begin{tabular}{lccc}
        \toprule
        Components & AP (\%)$\uparrow$ & SAP (\%)$\uparrow$ & PCE$\downarrow$ \\ 
        \midrule
        ControlNet  & 39.52 & 69.19 & 1.54 \\
        +PGI        & 47.80 & 70.63 & 1.51 \\
        +$\mathcal{L}_p$  & 43.81 & 70.37 & 1.48 \\
        +PGI+$\mathcal{L}_p$ & \textbf{49.50} & \textbf{70.84} & \textbf{1.43} \\
        \bottomrule
    \end{tabular}
\end{table}
\tabcolsep=0.12cm
\begin{table}[t]
    \centering
    \caption{{Results of pose guidance in upper body and full body on the cartoon subset from Human-Art dataset.} }
    \label{tab:upper}
    \begin{tabular}{ccccc}
        \toprule
         Scenes & Methods & AP (\%)$\uparrow$ & SAP (\%)$\uparrow$ & PCE$\downarrow$ \\ 
        \midrule
        \multirow{2}{*}{\makecell{Upper \\ Body}} & ControlNet  & 15.51 & 66.45 &0.63 \\
        & Ours & \textbf{31.87} & \textbf{70.05} &\textbf{0.57} \\
        \midrule
        
        \multirow{2}{*}{\makecell{Full \\ Body}} & ControlNet  & 30.78 & 70.12 &1.29 \\
        & Ours & \textbf{39.19} & \textbf{71.03} &\textbf{1.17} \\
        
        \bottomrule
    \end{tabular}
\end{table}

\begin{figure*}[h]
\centerline{\includegraphics[width=0.9\linewidth]{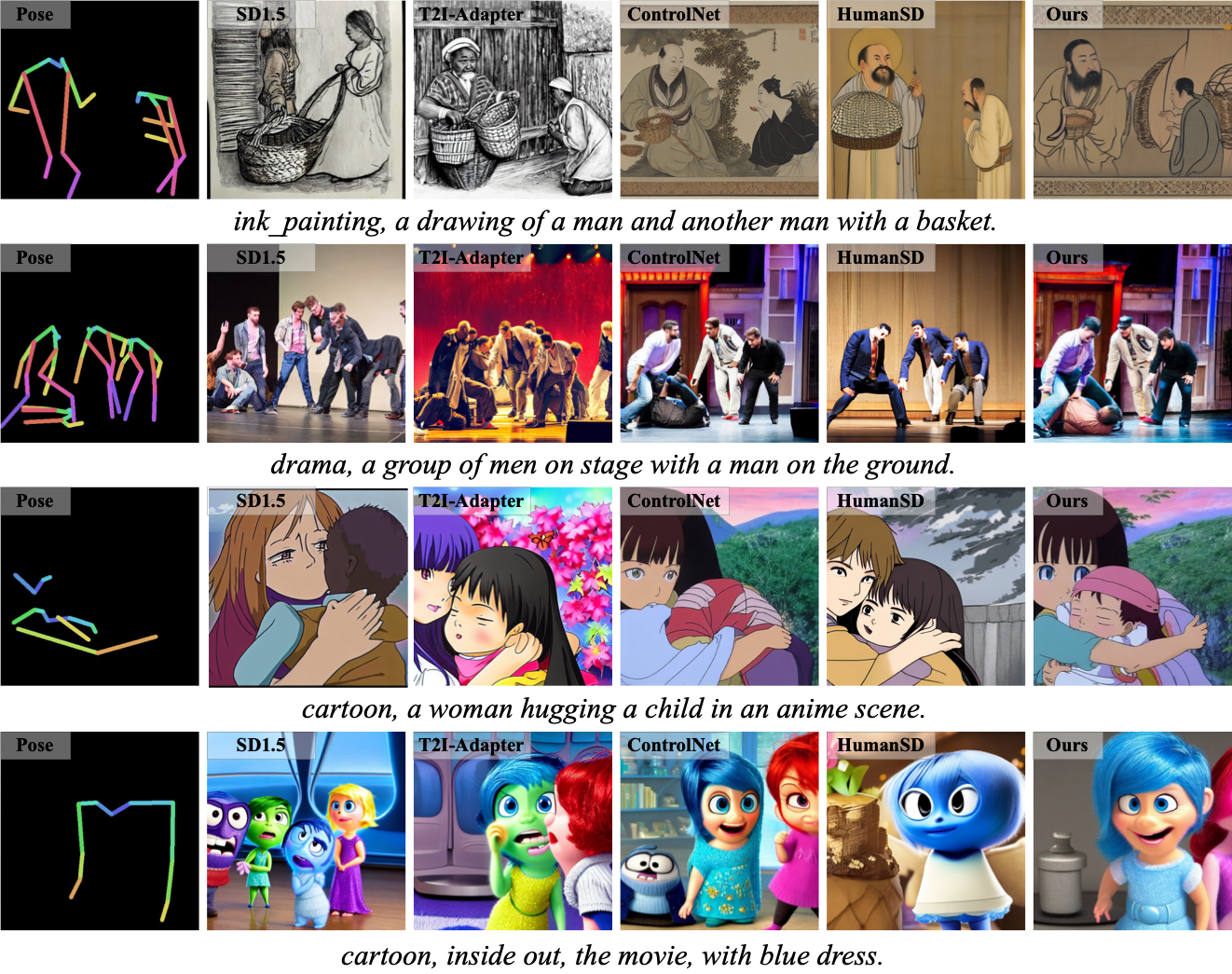}}
\caption{\textbf{Qualitative comparison between ours and other methods.} The samples are from the Human-Art dataset, where each row illustrates a sample along with its corresponding pose and prompt. } \label{Compa}
\end{figure*}

\tabcolsep=0.35cm
\begin{table}[h]
    \centering
    \caption{\textbf{Effects of different graph number.} The use of graph convolution layers at different stages significantly improves the quality of pose alignment. }
    \label{tab:graph-number}
    \begin{tabular}{c|c|c|c}
        \toprule
        \#. Graph & AP (\%)$\uparrow$ & SAP (\%)$\uparrow$ & PCE$\downarrow$ \\ 
        \hline
        0    & 41.85 & 69.96 & 1.57 \\
        1    & 42.75 &70.56 &1.52 \\
        2  & 46.59 & 70.59 & 1.58 \\
        3 & \textbf{47.80} & \textbf{70.63} & \textbf{1.51} \\
        \bottomrule
    \end{tabular}
\end{table}

\begin{table}[t]
    \centering
    \caption{\textbf{Results of different loss weights $\alpha$.} When $\alpha$ was set to 0.01, our model yielded the best performance.  }
    \label{tab:loss-weights}
    \begin{tabular}{c|c|c|c}
        \toprule
        $\alpha$ & AP (\%)$\uparrow$ & SAP (\%)$\uparrow$ & PCE$\downarrow$ \\ 
        \hline
        0.1    & 45.58 & 70.36 & 1.52 \\
        0.01   &\textbf{49.50} & \textbf{70.84} & \textbf{1.43} \\
        0.05 & 48.44 & 70.68 & 1.45 \\
        \bottomrule
    \end{tabular}
\end{table}

\subsection{Ablation Studies}


\subsubsection{All Components.}

We conducted experiments on the Human-Art dataset to validate the effectiveness of each component, as shown in Table \ref{tab:ablation}. Compared with ControlNet, the usage of the Pose Perception Loss \(\mathcal{L}_p\) led to an increase of 4.29\% in AP and 1.18\% in SAP, demonstrating that the pose perception loss effectively constrained the pose differences in the feature space from an alternative perspective. Incorporating the PGI further improved the performance of our model, with AP and SAP increasing by 8.28\% and 1.44\% respectively. This improvement indicated that PGI, benefited from its modeling of complex graph topological relationships, accurately captured high-order relationships between pose parts. It is evident that after PGI associates the information of different parts of the pose, our model produced significant improvements in both AP and SAP, addressing previous issues such as blurred limbs and unrealistic poses, as shown in Figure~\ref{multi-pose}. 

To verify the ability of portrait generation, we conducted additional experiments on \textit{upper body} and \textit{full body} samples from the cartoon subset in Human-Art, as shown in Table~\ref{tab:upper}. In the upper body scenario, our model outperformed ControlNet by 16.36\% in the AP. In the full body scenario it was 8.41\% higher in the AP. The results showed that compared with ControlNet, our model performs much better in upper body generation than in full body generation, further verifying the effectiveness of our PGI associations. The performance in the upper body scenario indicated that our model achieved good results in portrait generation, which demonstrates the high potential in many downstream applications.

\subsubsection{Graph Stages.}

We conducted empirical exploration to assess the performance of our model with varying numbers of graph structures, 
as shown in Table~\ref{tab:graph-number}. The results indicated that we can observe a significant performance increase when the number of graph layers increased from 1 to 4. We did not report more results due to computational resource limitations. In addition, we visualize our results of pose alignment with and without PGI in Figure~\ref{PGI_Com}. In the first row of samples, the leg pose showed a noticeable error and was partially missing without PGI. In the second row, the arm was in an incorrect pose. With PGI, pose guidance significantly improved image quality, bringing the generated pose of the image closer to the origin. This shows that capturing high-order relationships through the graph structure effectively facilitates information propagation between different parts of the pose.

\subsubsection{Pose Perception Loss.}

We investigated the impact of different loss weights $\alpha$ on Pose Perception Loss $\mathcal{L}_p$, as shown in Table~\ref{tab:loss-weights}. It can be found that the optimal loss weight is $0.01$. However, when the weight was set to 0.1, there was a significant performance decrease in AP, suggesting that it imposed over-high constraints on the diffusion loss. A higher weight for pose perception loss may lead to a failure of pose perception.

\section{Conclusion}

In this paper, we proposed a framework named Graph Relation Pose (GRPose), for pose-guided human image generation. We designed a Progressive Graph Integrator (PGI), which adopts a hierarchical structure within the Adapter to ensure that higher-order associations gradually correct the pose. Additionally, we proposed a novel pose perception loss, which uses a pre-trained pose estimation network to constrain the network, further enhancing the performance of pose alignment. In the future, we will try to extend our solution for 3D generation tasks~\cite{di2024hyper,luo2024trame}.

\section{Acknowledgments}
We acknowledge the support of the advanced computing resources provided by the Supercomputing Center of the USTC.



\bibliography{aaai25}

\begin{thebibliography}{36}
\providecommand{\natexlab}[1]{#1}

\bibitem[{Bhunia et~al.(2023)Bhunia, Khan, Cholakkal et~al.}]{bhunia2023person}
Bhunia, A.~K.; Khan, S.; Cholakkal, H.; et~al. 2023.
\newblock Person image synthesis via denoising diffusion model.
\newblock In \emph{CVPR}, 5968--5976.

\bibitem[{Bi{\'n}kowski et~al.(2018)Bi{\'n}kowski, Sutherland, Arbel, and Gretton}]{KID}
Bi{\'n}kowski, M.; Sutherland, D.~J.; Arbel, M.; and Gretton, A. 2018.
\newblock Demystifying mmd gans.
\newblock \emph{arXiv preprint arXiv:1801.01401}.

\bibitem[{Cheong, Mustafa, and Gilbert(2022)}]{cheong2022kpe}
Cheong, S.~Y.; Mustafa, A.; and Gilbert, A. 2022.
\newblock Kpe: Keypoint pose encoding for transformer-based image generation.
\newblock \emph{arXiv preprint arXiv:2203.04907}.

\bibitem[{Corona et~al.(2024)Corona, Zanfir, Bazavan, Kolotouros, Alldieck, and Sminchisescu}]{corona2024vlogger}
Corona, E.; Zanfir, A.; Bazavan, E.~G.; Kolotouros, N.; Alldieck, T.; and Sminchisescu, C. 2024.
\newblock VLOGGER: Multimodal diffusion for embodied avatar synthesis.
\newblock \emph{arXiv preprint arXiv:2403.08764}.

\bibitem[{Di et~al.(2024)Di, Yang, Luo, Xue, Chen, Yang, and Gao}]{di2024hyper}
Di, D.; Yang, J.; Luo, C.; Xue, Z.; Chen, W.; Yang, X.; and Gao, Y. 2024.
\newblock Hyper-3DG: Text-to-3D Gaussian Generation via Hypergraph.
\newblock \emph{IJCV}.

\bibitem[{Fan et~al.(2022)Fan, Sowmya, Meijering, and Song}]{fan2022fast}
Fan, L.; Sowmya, A.; Meijering, E.; and Song, Y. 2022.
\newblock Fast FF-to-FFPE whole slide image translation via Laplacian pyramid and contrastive learning.
\newblock In \emph{MICCAI}, 409--419. Springer.

\bibitem[{Goodfellow et~al.(2020)Goodfellow, Pouget-Abadie, Mirza, Xu, Warde-Farley, Ozair, Courville, and Bengio}]{goodfellow2020generative}
Goodfellow, I.; Pouget-Abadie, J.; Mirza, M.; Xu, B.; Warde-Farley, D.; Ozair, S.; Courville, A.; and Bengio, Y. 2020.
\newblock Generative adversarial networks.
\newblock \emph{Communications of the ACM}, 139--144.

\bibitem[{Han et~al.(2022)Han, Wang, Guo, Tang, and Wu}]{han2022vision}
Han, K.; Wang, Y.; Guo, J.; Tang, Y.; and Wu, E. 2022.
\newblock Vision gnn: An image is worth graph of nodes.
\newblock \emph{NeurIPS}, 35: 8291--8303.

\bibitem[{Heusel et~al.(2017)Heusel, Ramsauer, Unterthiner, Nessler, and Hochreiter}]{FID}
Heusel, M.; Ramsauer, H.; Unterthiner, T.; Nessler, B.; and Hochreiter, S. 2017.
\newblock GANs Trained by a Two Time-Scale Update Rule Converge to a Local Nash Equilibrium.
\newblock \emph{NeurIPS}.

\bibitem[{Ho and Salimans(2022)}]{ho2022classifier}
Ho, J.; and Salimans, T. 2022.
\newblock Classifier-free diffusion guidance.
\newblock \emph{arXiv preprint arXiv:2207.12598}.

\bibitem[{Ju et~al.(2023{\natexlab{a}})Ju, Zeng, Wang, Xu, and Zhang}]{ju2023human}
Ju, X.; Zeng, A.; Wang, J.; Xu, Q.; and Zhang, L. 2023{\natexlab{a}}.
\newblock Human-art: A versatile human-centric dataset bridging natural and artificial scenes.
\newblock In \emph{CVPR}, 618--629.

\bibitem[{Ju et~al.(2023{\natexlab{b}})Ju, Zeng, Zhao, Wang, Zhang, and Xu}]{humansd}
Ju, X.; Zeng, A.; Zhao, C.; Wang, J.; Zhang, L.; and Xu, Q. 2023{\natexlab{b}}.
\newblock Humansd: A native skeleton-guided diffusion model for human image generation.
\newblock In \emph{ICCV}, 15988--15998.

\bibitem[{Ju et~al.(2023{\natexlab{c}})Ju, Zeng, Zhao, Wang, Zhang, and Xu}]{ju2023humansd}
Ju, X.; Zeng, A.; Zhao, C.; Wang, J.; Zhang, L.; and Xu, Q. 2023{\natexlab{c}}.
\newblock Humansd: A native skeleton-guided diffusion model for human image generation.
\newblock In \emph{ICCV}, 15988--15998.

\bibitem[{Kingma and Ba(2014)}]{Kingma_Ba_2014}
Kingma, D.; and Ba, J. 2014.
\newblock Adam: A Method for Stochastic Optimization.
\newblock \emph{arXiv: Learning}.

\bibitem[{Kingma and Welling(2013)}]{vae}
Kingma, D.~P.; and Welling, M. 2013.
\newblock Auto-encoding variational bayes.
\newblock \emph{arXiv preprint arXiv:1312.6114}.

\bibitem[{Kipf and Welling(2016)}]{kipf2016semi}
Kipf, T.~N.; and Welling, M. 2016.
\newblock Semi-supervised classification with graph convolutional networks.
\newblock \emph{arXiv preprint arXiv:1609.02907}.

\bibitem[{Liang et~al.(2024)Liang, Hu, Huang, Di, Su, and Fan}]{liang2024tocoad}
Liang, Y.; Hu, Z.; Huang, J.; Di, D.; Su, A.; and Fan, L. 2024.
\newblock ToCoAD: Two-Stage Contrastive Learning for Industrial Anomaly Detection.
\newblock \emph{IEEE TIM}.

\bibitem[{Luo et~al.(2025)Luo, Di, Yang, Ma, Xue, Wei, and Liu}]{luo2024trame}
Luo, C.; Di, D.; Yang, X.; Ma, Y.; Xue, Z.; Wei, C.; and Liu, Y. 2025.
\newblock TrAME: Trajectory-Anchored Multi-View Editing for Text-Guided 3D Gaussian Splatting Manipulation.
\newblock \emph{IEEE TMM}.

\bibitem[{Lv et~al.(2021)Lv, Li, Li, Li, Lin, He, and Zuo}]{lv2021learning}
Lv, Z.; Li, X.; Li, X.; Li, F.; Lin, T.; He, D.; and Zuo, W. 2021.
\newblock Learning semantic person image generation by region-adaptive normalization.
\newblock In \emph{CVPR}, 10806--10815.

\bibitem[{Ma et~al.(2017)Ma, Jia, Sun, Schiele, Tuytelaars, and Van~Gool}]{ma2017pose}
Ma, L.; Jia, X.; Sun, Q.; Schiele, B.; Tuytelaars, T.; and Van~Gool, L. 2017.
\newblock Pose guided person image generation.
\newblock \emph{NeurIPS}, 30.

\bibitem[{Ma et~al.(2021)Ma, Peng, Wang, and Dong}]{ma2021must}
Ma, T.; Peng, B.; Wang, W.; and Dong, J. 2021.
\newblock Must-gan: Multi-level statistics transfer for self-driven person image generation.
\newblock In \emph{CVPR}, 13622--13631.

\bibitem[{Men et~al.(2020)Men, Mao, Jiang, Ma, and Lian}]{men2020controllable}
Men, Y.; Mao, Y.; Jiang, Y.; Ma, W.-Y.; and Lian, Z. 2020.
\newblock Controllable person image synthesis with attribute-decomposed gan.
\newblock In \emph{CVPR}, 5084--5093.

\bibitem[{Mou et~al.(2024)Mou, Wang, Xie, Wu, Zhang, Qi, and Shan}]{t2iadapter}
Mou, C.; Wang, X.; Xie, L.; Wu, Y.; Zhang, J.; Qi, Z.; and Shan, Y. 2024.
\newblock T2i-adapter: Learning adapters to dig out more controllable ability for text-to-image diffusion models.
\newblock In \emph{AAAI}, 4296--4304.

\bibitem[{Pan et~al.(2024)Pan, Qin, Li, Xue, and Chen}]{pan2024synthesizing}
Pan, X.; Qin, P.; Li, Y.; Xue, H.; and Chen, W. 2024.
\newblock Synthesizing coherent story with auto-regressive latent diffusion models.
\newblock In \emph{WACV}, 2920--2930.

\bibitem[{Podell et~al.(2023)Podell, English, Lacey et~al.}]{sdxl}
Podell, D.; English, Z.; Lacey, K.; et~al. 2023.
\newblock Sdxl: Improving latent diffusion models for high-resolution image synthesis.
\newblock \emph{arXiv preprint arXiv:2307.01952}.

\bibitem[{Radford et~al.(2021)Radford, Kim, Hallacy et~al.}]{clip}
Radford, A.; Kim, J.~W.; Hallacy, C.; et~al. 2021.
\newblock Learning transferable visual models from natural language supervision.
\newblock In \emph{ICML}, 8748--8763. PMLR.

\bibitem[{Ramesh et~al.(2022)Ramesh, Dhariwal, Nichol, Chu, and Chen}]{ramesh2022hierarchical}
Ramesh, A.; Dhariwal, P.; Nichol, A.; Chu, C.; and Chen, M. 2022.
\newblock Hierarchical text-conditional image generation with clip latents.
\newblock \emph{arXiv preprint arXiv:2204.06125}.

\bibitem[{Rombach et~al.(2022)Rombach, Blattmann, Lorenz, Esser, and Ommer}]{ldm}
Rombach, R.; Blattmann, A.; Lorenz, D.; Esser, P.; and Ommer, B. 2022.
\newblock High-resolution image synthesis with latent diffusion models.
\newblock In \emph{CVPR}, 10684--10695.

\bibitem[{Shen et~al.(2023)Shen, Ye, Zhang, Wang, Han, and Yang}]{shen2023advancing}
Shen, F.; Ye, H.; Zhang, J.; Wang, C.; Han, X.; and Yang, W. 2023.
\newblock Advancing pose-guided image synthesis with progressive conditional diffusion models.
\newblock \emph{arXiv preprint arXiv:2310.06313}.

\bibitem[{Wang et~al.(2024{\natexlab{a}})Wang, Guo, Li, and Wang}]{wang2024eulermormer}
Wang, F.; Guo, D.; Li, K.; and Wang, M. 2024{\natexlab{a}}.
\newblock Eulermormer: Robust eulerian motion magnification via dynamic filtering within transformer.
\newblock In \emph{AAAI}, volume~38, 5345--5353.

\bibitem[{Wang et~al.(2024{\natexlab{b}})Wang, Ghahremani, Li, Ommer, and Wachinger}]{wang2024stable}
Wang, J.; Ghahremani, M.; Li, Y.; Ommer, B.; and Wachinger, C. 2024{\natexlab{b}}.
\newblock Stable-Pose: Leveraging Transformers for Pose-Guided Text-to-Image Generation.
\newblock \emph{arXiv preprint arXiv:2406.02485}.

\bibitem[{Yang et~al.(2021)Yang, Wang, Liu, Gao, Ren, Zhang, Wang, Ma, Hua, and Gao}]{yang2021towards}
Yang, L.; Wang, P.; Liu, C.; Gao, Z.; Ren, P.; Zhang, X.; Wang, S.; Ma, S.; Hua, X.; and Gao, W. 2021.
\newblock Towards fine-grained human pose transfer with detail replenishing network.
\newblock \emph{IEEE TIP}, 2422--2435.

\bibitem[{Yang et~al.(2023)Yang, Zhang, Song et~al.}]{yang2023diffusion}
Yang, L.; Zhang, Z.; Song, Y.; et~al. 2023.
\newblock Diffusion models: A comprehensive survey of methods and applications.
\newblock \emph{ACM Computing Surveys}, 1--39.

\bibitem[{Zhang, Rao, and Agrawala(2023)}]{controlnet}
Zhang, L.; Rao, A.; and Agrawala, M. 2023.
\newblock Adding conditional control to text-to-image diffusion models.
\newblock In \emph{ICCV}, 3836--3847.

\bibitem[{Zhang et~al.(2022)Zhang, Yang, Lai, and Xie}]{zhang2022exploring}
Zhang, P.; Yang, L.; Lai, J.-H.; and Xie, X. 2022.
\newblock Exploring dual-task correlation for pose guided person image generation.
\newblock In \emph{CVPR}, 7713--7722.

\bibitem[{Zhao et~al.(2024)Zhao, Chen, Chen et~al.}]{unicontrolnet}
Zhao, S.; Chen, D.; Chen, Y.-C.; et~al. 2024.
\newblock Uni-controlnet: All-in-one control to text-to-image diffusion models.
\newblock \emph{NeurIPS}, 36.

\end{thebibliography}

\end{document}